\documentclass[letterpaper, 10 pt, conference]{ieeeconf}  

\IEEEoverridecommandlockouts                              
\usepackage[pdftex]{graphicx}
\graphicspath{{../pdf/}{../jpeg/}}
\DeclareGraphicsExtensions{.pdf,.jpeg,.png}

\usepackage{url}
\usepackage{subfigure}
\usepackage{color}

\title{\LARGE \bf
Towards Quality of Service and Resource Aware Robotic Systems \\ 
through Model-Driven Software Development
}

\author{Andreas Steck and Christian Schlegel\\
University of Applied Sciences Ulm\\
        Department of Computer Science, Prittwitzstr. 10, 89075 Ulm, Germany\\
	email: {\tt\small \{steck, schlegel\}@hs-ulm.de}}

\begin{document}

\maketitle
\thispagestyle{empty}
\pagestyle{empty}

\begin{abstract}
Engineering the software development process in robotics is one of the basic
necessities towards industrial- strength service robotic systems. A major
challenge is to make the step from code-driven to model-driven systems. This is
essential to replace hand-crafted single-unit systems by systems composed out of
components with explicitly stated properties. Furthermore, this fosters reuse
by separating robotics knowledge from short-cycled implementational
technologies. Altogether, this is one but important step towards ``able''
robots.

This paper reports on a model-driven development process for robotic
systems. The process consists of a robotics meta-model with first explications
of \textit{non-functional properties}. A model-driven toolchain based on
\textit{Eclipse} provides the model transformation and
code generation steps. It also provides design time analysis of resource
parameters (e.g. schedulability analysis of realtime tasks) as a first step
towards overall \textit{resource awareness} in the development of integrated
robotic systems. The overall approach is underpinned by several real world
scenarios.
\end{abstract}

\section{Introduction}
Nowadays, implementing complete robotic systems is still more of an art than a
systematic engineering process. Integrating the various libraries currently is
more like plumbing. Essential properties are mostly hidden in the software
structures. In particular, \textit{non-functional properties} and
\textit{Quality of Service (QoS)} parameters are not explicated. Although, these
\textit{non-functional properties} are considered being mandatory for embodied
and deployed robots in real world, they are not yet addressed in a systematic
way in most robotics software development processes. Thus, these properties
cannot be taken into account during the design process, the system deployment
and the dynamically changing runtime configurations.

The relevance of \textit{non-functional properties} and \textit{QoS} parameters
arises from at least two observations of real world operation: (i) some
components have to guarantee \textit{QoS} (e.g. adequate response times to
ensure collision avoidance), (ii) the huge number of different behaviors needed
to handle real world contingencies (behaviors cannot all run at the same time
and one depends on situation and context aware configuration management and
resource assignment). Instead of allocating all resources statically, the system
should dynamically adapt itself. Thereby, appropriate \textit{QoS} parameters
and resource information are to be taken into account.

For example, if the current processor load does not allow to run the navigation
component at the highest quality level, the component should switch to a lower
quality level, resulting in a reduced navigation velocity, but still ensuring
safe navigation.

\section{Motivation} 
Service robots are expected to fulfill a whole variety of tasks under very
different conditions. We need various components providing the desired
capabilities to build a full-fledged robotic system. 
Even the software components should be \textit{COTS} components to foster reuse
and ensure maturity and robustness. This kind of composability inevitably
depends on assured services with explicated \textit{QoS} parameters and required
resources. That is even more important in service robotics where resources are
strictly limited. Nevertheless, these are per se not explicated and are thus not
accessible in most service robotic systems.

The desired approach should compose systems out of building blocks with assured
services, both enriched with \textit{QoS} parameters and resource information.

This would, for example, allow to perform realtime schedulability analysis,
performance analysis and would also allow to check whether response times of
services can be guaranteed.
Furthermore, this allows to check whether the desired mapping of components
to processors and communication busses provides enough computational and memory
resources and bandwidth.

Depending on the executed task and the current state of the environment, the
component interactions and configurations are different. Therefore, we cannot
simply design a robotic software system as static system by considering only a
small number of different configurations. In fact, we depend on dynamic system
configurations at runtime. That also requires to take resource limitations into
account.

Therefore, \textit{resource awareness} will be mandatory in robotics at all
phases of design, development, deployment and even at runtime.

As a consequence, one should establish a development process which can cope with
\textit{QoS} parameters and resource information at all development phases of
the entire system. Essential for such a process is to describe the system in an
abstract formal representation. \textit{Model-driven engineering
(MDE)} is the only known approach to make these relevant parameters explicit and
accessible. Models abstract from unnecessary details and give the developers a
domain-specific view on the system \cite{mdeComplex}.

Another significant benefit of \textit{model-driven software development} is the
decrease of development time and effort while increasing flexibility, reuse and
the overall system quality. Design patterns, best practices and approved
solutions can be made available within the code generators such that even
novices can immediately take advantage from that coded and immense experience.
A demanding challenge in making the step from code-driven to model-driven
systems is the definition of an appropriate meta-model.

\section{Related Work}
Driven by the avionics and automotive industries,
\textit{distributed realtime and embedded (DRE)} systems denote an established
research community, which already deals with questions relevant in robotics as
well.

For example, the {\em Artist2 Network of Excellence on Embedded Systems Design}
\cite{artist2} addresses highly relevant topics concerning realtime components
and realtime execution platforms. Furthermore, many symposia and conferences
directly tackle the problem of component architectures for embedded systems and
service oriented architectures in embedded systems \cite{isorc09}.

The automotive industry is trying to establish the \textit{AUTOSAR} standard
\cite{autosar} for software components and model-driven design of such
components. \textit{AUTOSAR} will provide a software infrastructure for
automotive systems, based on standardized interfaces. Related to
\textit{AUTOSAR}, the ongoing \textit{RT-Describe} project \cite{rtdescribe}
addresses resource aspects. To adapt the system during runtime,
\textit{RT-Describe} relies on self-description of the components. Software
components shall be enabled to autonomously reconfigure themselves, for example,
by deactivating functions that are currently not needed. \textit{RT-Describe}
models are based on \textit{UML} \cite{omgUml}, \textit{SysML} \cite{omgSysml}
and \textit{MARTE} \cite{omgMarte}.

The \textit{OMG MARTE} \cite{omgMarte} activity provides a standard for modeling
and analysis of realtime and embedded systems. They provide a huge number
of \textit{non-functional properties (NFPs)} to  describe both, the internals
and externals of the system. Mappings to analysis models (\textit{CHEDDAR},
\textit{RapidRMA}) are available\footnote{http://www.omgmarte.org/Tools.htm} to
perform scheduling analysis of \textit{MARTE} models. This part of the
\textit{MARTE} profile is of interest to the robotics community.

The major differences that arise in robotics compared to other domains are
\textit{not} the huge number of different sensors and actuators or the various
hardware platforms. Instead, the differences are the context and situation
dependant reconfigurations of interactions and a prioritized assignment of
limited resources to activities even at runtime -- again depending on context
and situation.
Thus, mastering the huge amount of different configurations in robotic systems
is far more complex than in common \textit{DRE} systems or cars.

Robotic frameworks \cite{rosta}, like \textit{Player/Stage} \cite{player},
\textit{ROS} \cite{ros} and \textit{MSRS} \cite{msrs} are widely in use and many
of them offer a rich support for common robotic hardware. 
All posses design tools, ranging from simple ones up to complex ones like visual
programming in \textit{Microsoft Robotics Developer Studio} \cite{msrs}. The
same holds true for middleware systems.
The \textit{OMG Data Distribution Service for Realtime Systems (DDS)}
\cite{omgdds} standard provides the concepts of a publish-subscribe middleware
structure with several integrated \textit{QoS} parameters. These tools and
libraries coexist, without any chance of interoperability. Each tool and
framework has attributes that favors its use. They all do not make the step
towards \textit{MDSD}. However, only this step provides the semantics to become
independent of specifics of equally suitable, but not easily replaceable,
frameworks. The essential benefit would be to finally enable productive use of
all those codebases due to a \textit{MDE} approach.

An introduction into robotics component-based software engineering (\textit{CBSE})
can be found in \cite{brugali2009,brugali2010}. Several important design principles and 
implementation guidelines that enable the development of reusable and maintainable software 
building-blocks are motivated in detail.

Within the robotics community, a model-based approach is meanwhile considered
valuable and the interest of framework developers becomes focused on
\textit{model-driven software development (MDSD)}.

The \textit{BRICS} project specifically aims at exploiting {\em MDE} as enabling approach 
to reducing the development effort of engineering robotic systems by making best practice  
robotic solutions more easily reusable. They plan to create a \textit{BRICS}
Integrated Development Environment, which will be based on Eclipse and will
provide the robotics community with an MDE toolchain. \cite{brics-ICR2010}

The \textit{OROCOS} \cite{orocos03} project conforms very well to our ideas.
They currently work on the integration of their framework into a model-driven
toolchain. Their focus is on hard realtime motion control applications.
\textit{OROCOS} provides \textit{hotspots} to be filled in by the component
developer.

{\em GenoM3} \cite{anthony2010} proposes scripting mechanisms to bind component
skeleton templates. This does not reach the abstraction level of meta-models and
severely restricts the application architecture to the given narrow template 
structure.

The \textit{OMG Robotics Domain Task Force} \cite{omg-robotics} develops a
standard to integrate robotic systems out of modular components. As in our work
the components comprise an internal state automaton and interact via service
ports. However, these ports are defined in a very generic way. The
concept of execution contexts decouple the business logic from the thread of
control. During the deployment the user has to assign appropriate execution
contexts to the components. In our approach the user does not necessarily need
any knowledge about the internal structure of the components during deployment.
But as the elements and parameters are made explicit they can be accessed by
analysis tools, for example. Unfortunately the reference implementation (OpenRTM)
requires user-code to directly interact with the PSI-level even for core model
parts, like communication. The user-code gets far to strong bound to middleware
(\textit{CORBA}) specifics.

An interesting approach which is similar to our work is presented in
\cite{joserV3CMM}. The \textit{3-View Component Meta-Model (V$^{3}$CMM)}
comprises three complementary views (\textit{structural view},
\textit{coordination view} and \textit{algorithmic view}) to model
component-based systems independent of the execution platform. 
\textit{V$^{3}$CMM} encourages describing the component's algorithms (as an
inherent part of the systems model) in a manner which is similar to \textit{UML
activity diagrams}. 
In comparison to that, we model the component hull ``only''. Inside of our
components, the developers can integrate their algorithms and libraries
(\textit{OpenCV}, \textit{Qt}, \textit{OpenRAVE}, etc.) without any
restrictions. In \textit{V$^{3}$CMM} the components interact by calling the
mutually provided interface operations. In our meta-model we provide strictly
enforced interaction patterns to decouple the sphere of influence and thus
partition the overall complexity of the system. This ensures reusability and
interchangeability of the components.

It is worth highlighting that we focus on defining an abstract
meta-model ensuring the modeling and analysis of robotic systems. Hence,
we extend our current component model iteratively and express it in an abstract
way (\textsc{SmartMARS} meta-model) underpinned by different but replaceable
implementational technologies.

\section{Model-Driven Software Development for Robotic Systems}
Crucial to make the step from code-driven to model-driven engineering is to
define a meta-model and a matured \textit{model-driven software development}
process which considers the properties explicated in the models. The challenge
is to identify appropriate levels of abstraction and to handle the properties,
attached to the elements of the model, in these different levels. As a result of
our work, we propose a development process and a meta-model for robotic systems.

\begin{figure}
 \centering
 \includegraphics[width=0.4\textwidth,keepaspectratio=true]
{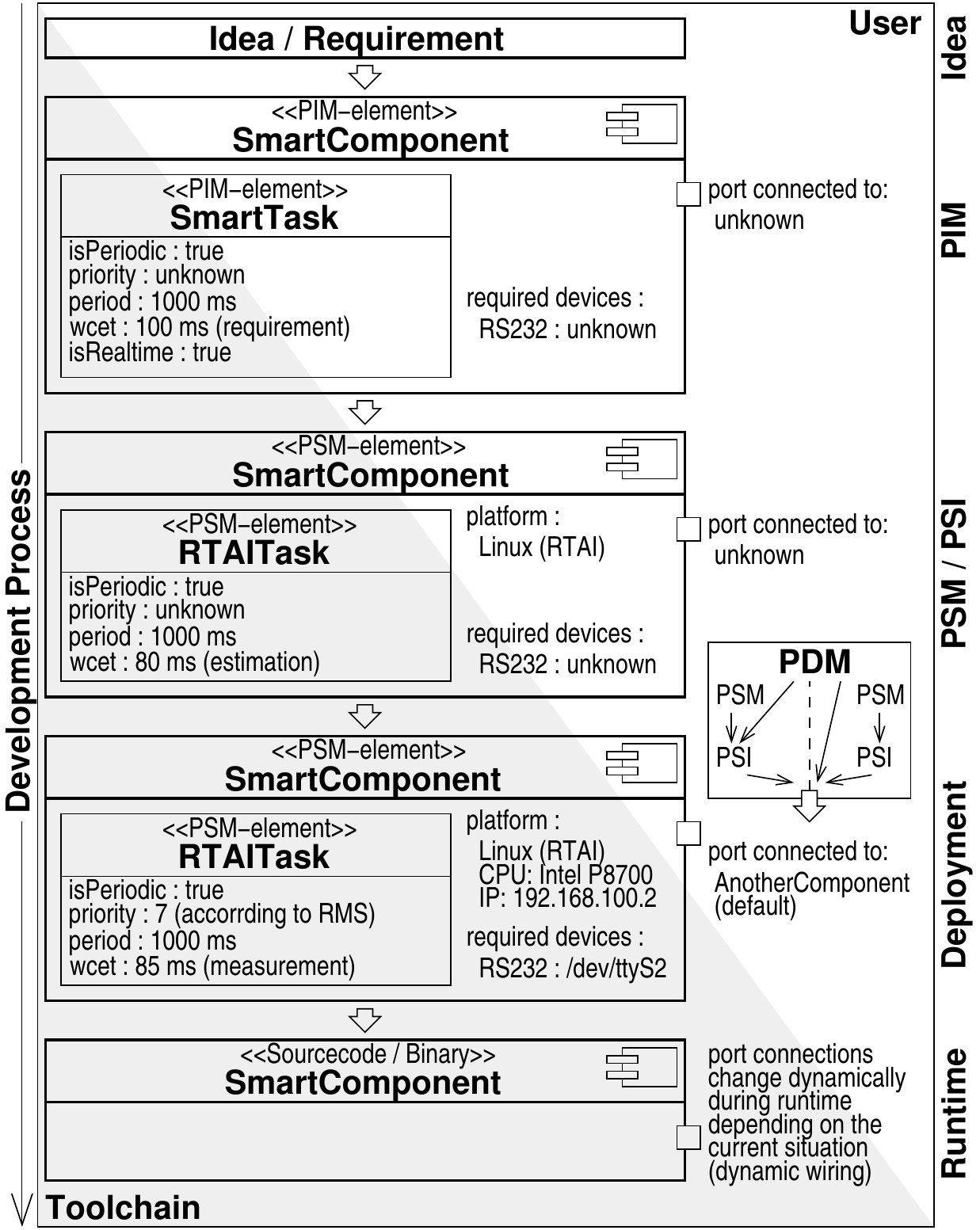}
 \caption{The development process at-a-glance.}
 \label{fig:devProcess}
\end{figure}
\subsection{The Development Process}
The development process (fig. \ref{fig:devProcess}) aims \textit{not} to
exhaustively model a robotic system at all views such that afterwards one has
just to push the button of a toolchain and to await the executables. We are
convinced that this will remain unrealistic. However, there is already a huge
benefit if such a process can help to master the complexity of our systems and
in particular, allows to manage and explicate \textit{non-functional
properties}. Starting with an idea, the model will be enriched during
development time until it finally gets executable in form of deployed software
components.

The major steps are, firstly, to describe the system in a model-based
representation (\textit{platform independet model - PIM}) which is independent of
the underlying framework, middleware structures, operating system and programming
language. In this design phase, several system properties are either unknown or
only known as requirements.

The second step is to transform the \textit{PIM} into a \textit{platform specific
model (PSM)}. After model checks, the platform independent elements are
transformed into the appropriate elements of the platform specific meta-model.
These elements represent the characteristics of the underlying environment
(middleware structures, operating system, framework, etc.). The \textit{PSM} is
enriched with properties which are known due to the knowledge about the platform
specific characteristics. However, some parameters can still be unknown and
are added not until the deployment of the component. Furthermore, the
\textit{PSM} is transformed into the platform specific implementation
(\textit{PSI}). The developers add their algorithms and libraries
(\textit{user-code}) with the guidance of the toolchain. The user code is
protected from modifications made by the code generator due to the \textit{generation gap
pattern} \cite{generationGap} which is based on inheritance.

The third step is to deploy the components to the different platforms
of the robotic system. The capabilities and characteristics of the target
platforms are defined by the \textit{platform description model (PDM)}. In this
phase, the model is enriched by the knowledge about the target platforms.
Further model checkings are performed to verify the constraints attached to the
components against the capabilities of the system (e.g. is executable only on a
certain type of hardware, needs one serial port, etc.). The \textit{QoS}
parameters of the interaction patterns, for example, can be cross-checked
whether they can be satisfied. Furthermore, special analysis models can be
generated out of the deployment model. These special models can be used to
perform realtime schedulability analysis, for example.

The fourth step is to run the system according to the deployment model. Certain
properties can still be unknown and will be reasoned during runtime.
Additionally, the wiring between the components can change during
runtime depending on the current situation of the robot. Even if it would be
possible to calculate the required resources for a worst-case and
rare-event situation in advance, it would not be possible and efficient to
provide these resources on a standard service robot. Dynamic adaptions and
\textit{resource awareness} are mandatory to be able to fulfill a variety of
different tasks even with limited overall resources. 

Absolutely essential for the development process is the definition of
an abstract meta-model to describe robotic systems independently of the
implementational technology.

\subsection{The \textsc{SmartMARS} meta-model}
\begin{figure}
 \centering
 \includegraphics[width=0.43\textwidth,keepaspectratio=true]
{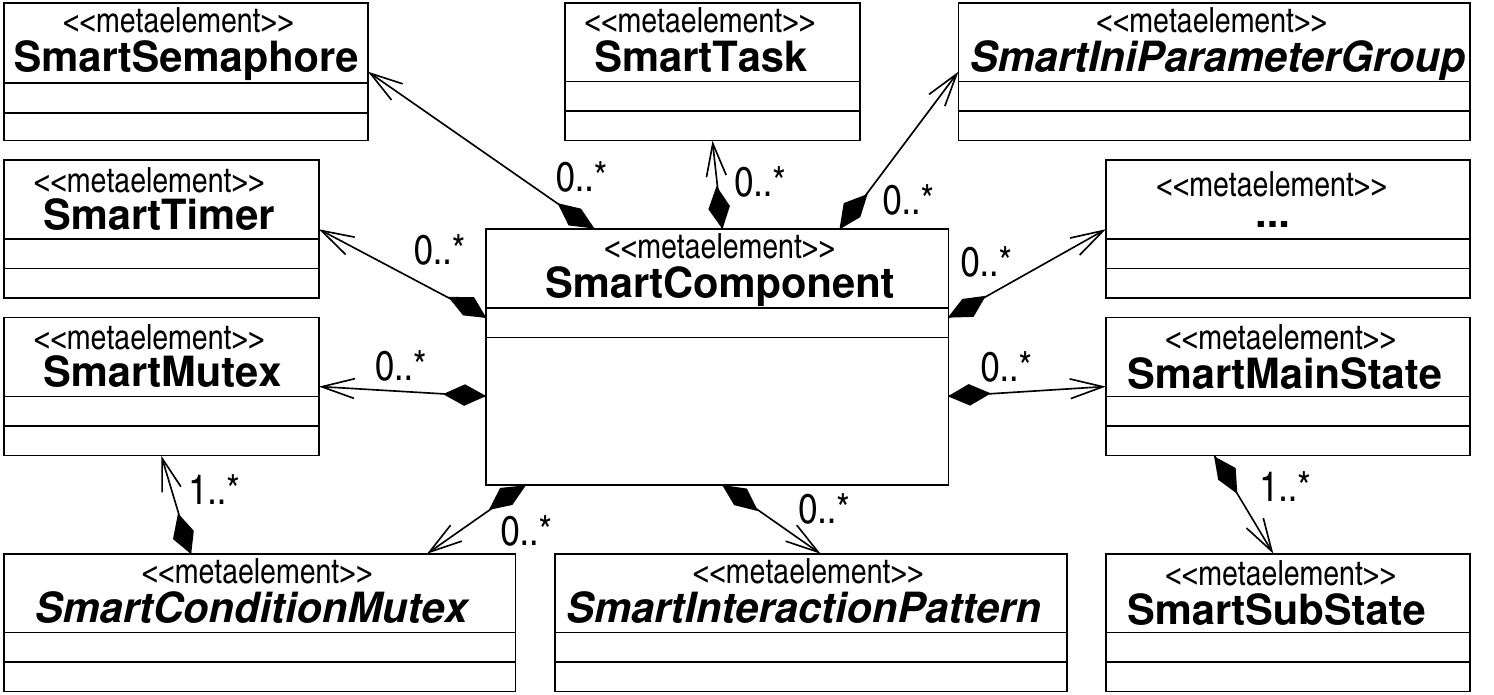}
 \caption{Excerpt of the \textsc{SmartMARS} meta-model.}
 \label{fig:metamodel}
\end{figure}

As a promising starting point to define the meta-model, we adapt and adopt
well-established concepts of the \textsc{SmartSoft} framework
\cite{schlegel2006}, which has been continuously extended and used in building
robotic systems for more than a decade now. The \textsc{SmartSoft} concepts are
independent of the implementation technology and scale from 8-bit
microcontrollers up to large scale systems \cite{Schlegel2009}. Two reference
implementations are available on sourceforge \cite{smartsoft-sourceforge}. The
first implementation is based on \textit{CORBA} and the second one on
\textit{ACE} \cite{ace} only.

\begin{figure*}
\begin{minipage}{0.85\textwidth}
 \includegraphics[width=0.99\textwidth,keepaspectratio=true]{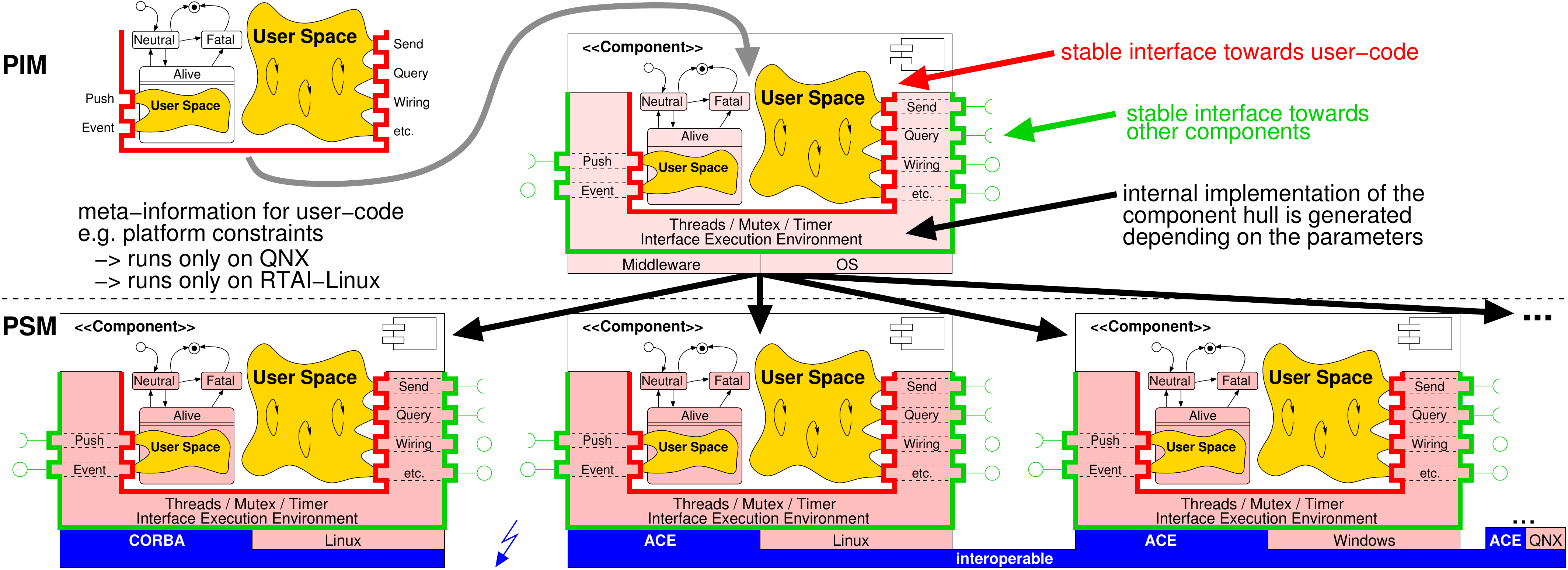}
\end{minipage}
\begin{minipage}{0.14\textwidth}
 \includegraphics[width=1.0\textwidth,keepaspectratio=true]{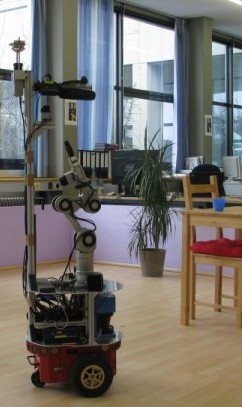}
\end{minipage}
 \caption{\textit{Left:} Refinement steps of component development.
 \textit{Right:} Our robot Kate.}
 \label{fig:componentDev}
\end{figure*}

Consequently, we propose \textsc{SmartMARS} (Modeling and Analysis of Robotic
Systems) (fig. \ref{fig:metamodel}) as an abstract meta-model, which addresses
modeling and analysis of robotic systems. 
The basic concept behind the meta-model are loosely coupled components
offering/requiring services. These services are based on strictly enforced
interaction patterns (table \ref{tab:interactionpatterns}) that transmit
communication objects \cite{schlegel2006} \cite{schlegel-diss}.
\begin{table}
\caption{The set of interaction patterns.}
\label{tab:interactionpatterns}
\centering
\begin{tabular}{|l||l|l|}
\hline \textbf{Pattern}     & \textbf{Relationship}             & \textbf{Communication Mode} \\
\hline send        			& client/server 		& one-way communication \\
\hline query       			& client/server 		& two-way request \\
\hline push newest 			& publisher/subscriber	& 1-to-n distribution \\ 
\hline push timed 			& publisher/subscriber	& 1-to-n distribution \\
\hline event 				& client/server 		& asynchronous notification \\
\hline
\hline state				& master/slave			& activate/deactivate component \\
							& 						& services \\ 
\hline wiring      			& master/slave			& dynamic component wiring \\
\hline
\end{tabular}
\end{table}
They provide a precisely defined semantics and describe the outer view of a
component, independent of its internal implementation. The patterns decouple the
sphere of influence and thus partition the overall complexity of the system.
Internal characteristics can never span across components. The interaction
patterns provide stable interfaces towards the user code inside of the component
and towards the other components independent of the underlying middleware
structure (fig. \ref{fig:componentDev}). The interfaces can be used in completely
different access modalities as they are not only forwarding method calls
but are standalone entities. The \textit{query} pattern, for example, provides
both, synchronous and asynchronous access modalities at the client side and a
handler based interface at the server side. Interaction patterns are annotated
with \textit{QoS} parameters (e.g. cycle times for push timed pattern, timeouts
for query and event pattern).
Dynamic reconfiguration of the components at runtime is supported by a
\textit{param} port to send name-value pairs to the components, a \textit{state}
port to activate/deactivate component services and \textit{dynamic wiring} to
change the connections between the components.

The \textit{state pattern} is used by a component to manage transitions between service
activations, to support housekeeping activities (entry/exit actions) and to hide
details of private state relationships (appears as stateless interface to the
outside).

\textit{Dynamic wiring} is the basis for making both, the control flow and the
data flow configurable. This is required for situated and task dependent
composition of skills to behaviors.
 
\begin{figure}
 \centering
 \includegraphics[width=0.35\textwidth,keepaspectratio=true]
{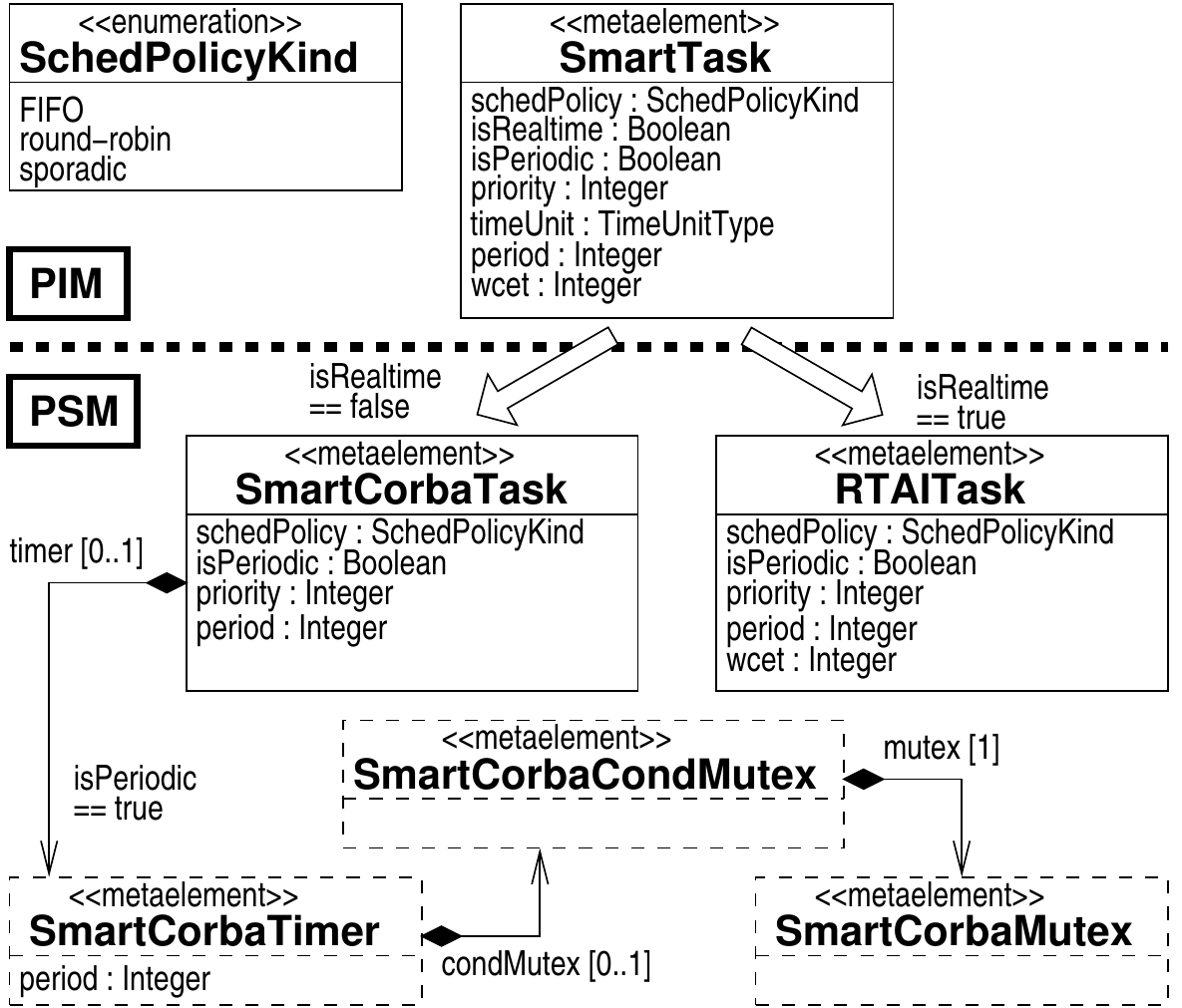}
 \caption{The transformation of the \textit{PIM} and the \textit{PSM} by means
 of the \textit{SmartTask} metaelement.}
 \label{fig:task}
\end{figure}

The \textit{wiring pattern} supports dynamic wiring of services from outside (and
inside) a component by exposing service requestors of a component as ports. The
wiring pattern allows to connect service requestors to service providers
dynamically at run time. A service requestor is connected only to a compatible
service provider (same pattern and communication object). Disconnecting a service
requestor automatically performs all housekeeping activities inside a
communication pattern to sort out not yet answered and pending calls, for
example, by iterating through the affected state automatons (inside interaction
patterns) and thus properly unblocking method calls that otherwise would never
return. For example, the wiring mechanism already properly sorts out effects of a
server being destroyed while clients are in the process of connecting to it. Of
course, this relieves a component builder from a huge source of potential
pitfalls.

\textsc{SmartMARS} covers two different views: (i) it is a completely abstract
meta-model for modeling and analysis of robotic systems and (ii) it is a
concrete reference implementation implemented as an \textit{UML profile}.

\section{The \textsc{SmartSoft} MDSD Toolchain}
As proof of concepts and to gain more experience on how to further investigate
on the development of \textsc{SmartMARS}, we designed, implemented and provided
the \textsc{SmartSoft MDSD Toolchain} \cite{smartsoft-sourceforge} which is
based on the Eclipse Modeling Project \cite{eclipse}.

\subsection{Mapping of Abstract Concepts}

Fig. \ref{fig:task} shows the transformation of the \textit{PIM} and the
\textit{PSM} by the example of the \textit{SmartTask} metaelement and the
\textit{CORBA} based \textit{PSM}. The \textit{SmartTask} (\textit{PIM})
comprises several parameters which are necessary to describe the task behavior
and its characteristics independent of the implementation.

Depending on the \texttt{isRealtime} attribute and the capabilities of the target
platform (\textit{PDM}) the \textit{SmartTask} is either mapped onto a
\textit{RTAITask} or a non-realtime \textit{SmartCorbaTask}. If hard realtime
capabilities are required and are not offered by the target platform, the toolchain 
reports this missing property. To perform realtime schedulability tests, the
attributes (\texttt{wcet, period}) of the \textit{RTAITasks} can be forwarded
to appropriate analysis tools. 

In case the attributes specify a \texttt{non-realtime}, \texttt{periodic}
\textit{SmartTask}, the toolchain extends the \textit{PSM} by the elements
needed to emulate periodic tasks (as this feature is not covered by standard tasks).

In each case the user integrates his algorithms and libraries into the stable
interface provided by the \texttt{SmartTask} (user view) independent of the
hidden internal mapping of the \texttt{SmartTask} (generated code).

\subsection{Development of Components}

\begin{figure}
 \centering
 \includegraphics[width=0.65\columnwidth,keepaspectratio=true]
{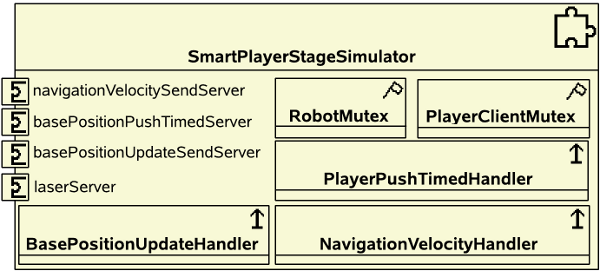}
 \caption{Model of the SmartPlayerStageSimulator component modeled with the
 \textsc{SmartSoft MDSD Toolchain}}
 \label{fig:playerComp}
\end{figure}

The major steps to develop a \textsc{SmartSoft} component are depicted in figure
\ref{fig:componentDev} on the left. The component developer models the component
in a platform independent representation using the \textsc{SmartMARS} \textit{UML
profile}. He focuses on the component hull, which comprises, for example, service
ports and tasks -- without any implementation details in mind (fig.
\ref{fig:playerComp}). Pushing the button, the toolchain verifies the model
(\textit{oAW check}), transforms it into an appropriate \textit{PSM} (\textit{oAW
xTend}) and generates the \textit{PSI} (\textit{oAW xPand}). Accordingly, the
developer integrates his algorithms and libraries without any restrictions, but
with the guidance of the toolchain (\textit{oAW recipes}). The \textit{oAW
recipes} assist, for example, the handling and usage of the interaction patterns
in the user part of the source code. Tags are used to indicate user code
constraints (e.g. runs only on RTAI-Linux) and need to be set in the model by the
user.

\begin{figure}
\begin{minipage}{0.3\columnwidth}
 \centering
 \includegraphics[width=1\columnwidth]
{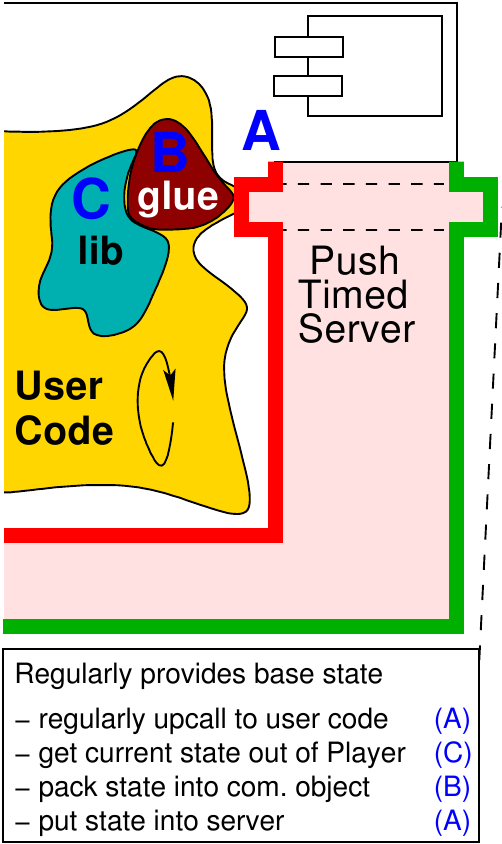}
\end{minipage}
\begin{minipage}{0.69\columnwidth}
 \centering
 \includegraphics[width=1\columnwidth]
{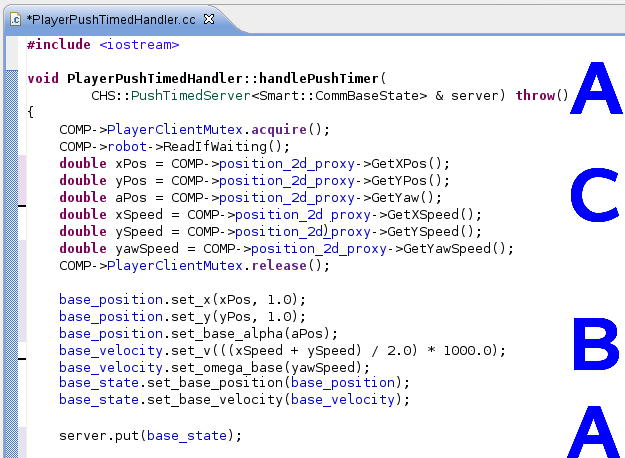}
\end{minipage}
 \caption{The service port regularly provides a base state (pose, velocity etc.) by means of
    the {\em push timed} pattern.}
 \label{fig:library-integration}
\end{figure}

Fig. \ref{fig:library-integration} illustrates how existing libraries are easily integrated
into the user space of the component hull and how the glue logic looks like to link existing
libraries (or code generated by other tools) to the generated component hull.

\subsection{Deployment of Components}

To create a system deployment the developer imports the components needed for the
scenario into the deployment model. He maps the components onto the desired
target platform and defines, for example, the initial wiring between the
components.

\section{Example}
\begin{figure}
 \centering
 \includegraphics[width=0.49\textwidth,keepaspectratio=true]{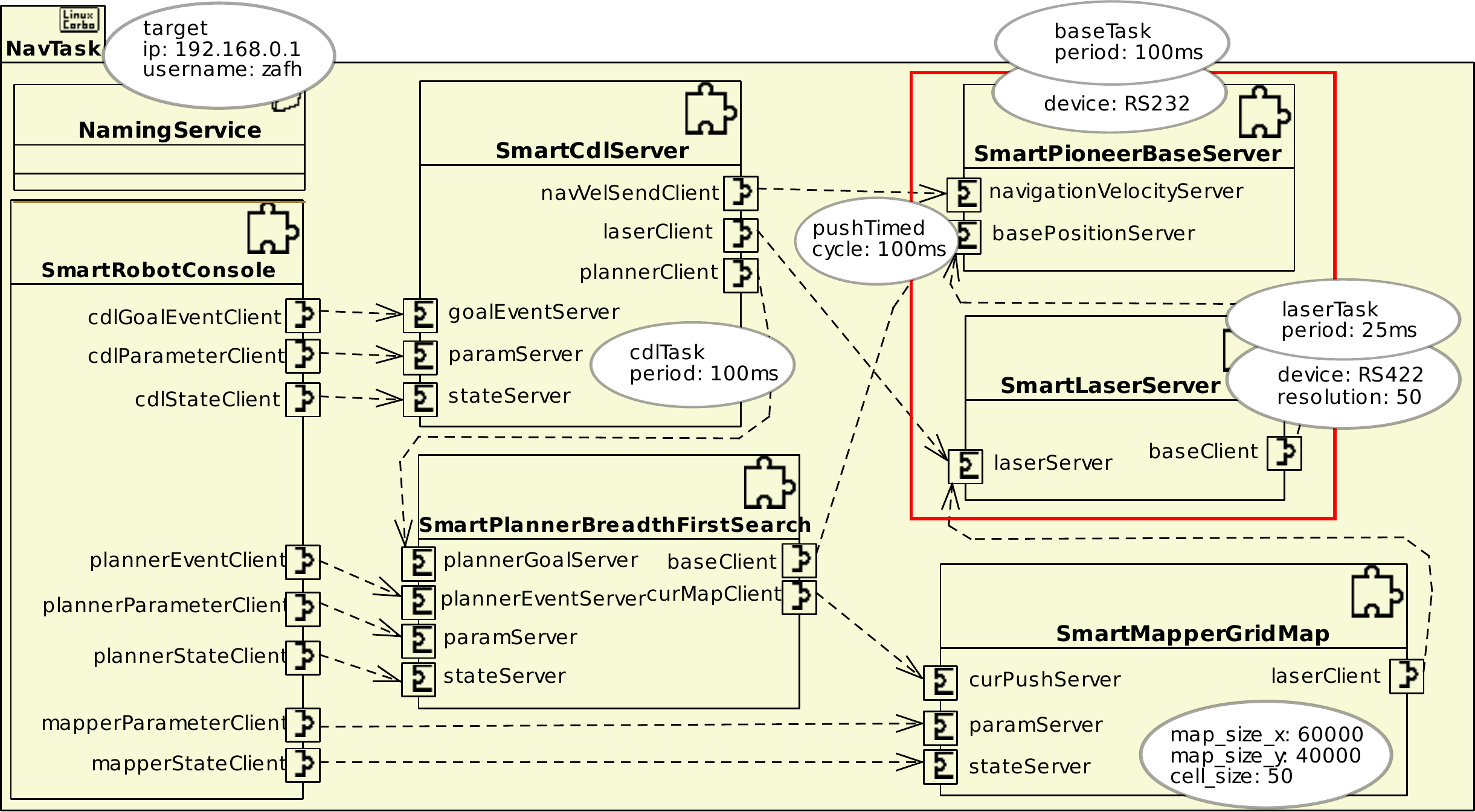}
 \caption{Deployment diagram of a navigation task. Switching to simulation is
  done by replacing services (e.g. the Player/Stage component provides services
  of P3DX and LRF). Ellipses show selected non-functional properties defined in
  the models.}
 \label{fig:navTask}
\end{figure}

The model-driven toolchain has been used to build numerous robotic components (navigation,
manipulation, speech, person detection and recognition, simulators) reusing many 
already existing libraries within the component hulls ({\em OpenRAVE, Player/Stage, GMapping, MRPT,
Loquendo, Veri-Look, OpenCV}, etc.). These components are reused in different
arrangements (fig. \ref{fig:navTask}) to implement, for example, {\em Robocup@Home} scenarios ({\em Follow Me}, {\em Shopping Mall}, 
{\em Who-is-Who}).

The \textit{non-functional properties} specified in the different modeling
levels enable analysis of resource usage and verification of resource
constraints. Hard realtime schedulability analysis, for example, is performed by
a model-to-model transformation from a deployment model into a \textit{CHEDDAR}
\cite{cheddar} specific analysis model.

\section{Conclusion and Future Work}
The proposed development process and meta-model allows the explication,
management and analysis of \textit{non-functional properties}. We made
those parameters accessible during development and deployment to check for
guarantees and resource awareness in a systematic way. The next steps are to
extend the \textsc{SmartMARS} meta-model and to further make use of it for
analysis, verification and simulation at design-time as well as at run-time.

\section{ACKNOWLEDGMENTS}
This work has been conducted within the {\em ZAFH Service\-robotik} 
(http://www.servicerobotik.de/). The authors gratefully acknowledge the 
research grants of state of Baden-W{\"u}rttemberg and the European Union.

We thank Dennis Stampfer for his extraordinary support in implementing the
\textsc{SmartSoft MDSD Toolchain}.\\

\bibliographystyle{IEEEtran}
\bibliography{IEEEabrv,2010_iros_dslrob_steck}

\begin{thebibliography}{10}
\providecommand{\url}[1]{#1}
\csname url@rmstyle\endcsname
\providecommand{\newblock}{\relax}
\providecommand{\bibinfo}[2]{#2}
\providecommand\BIBentrySTDinterwordspacing{\spaceskip=0pt\relax}
\providecommand\BIBentryALTinterwordstretchfactor{4}
\providecommand\BIBentryALTinterwordspacing{\spaceskip=\fontdimen2\font plus
\BIBentryALTinterwordstretchfactor\fontdimen3\font minus
  \fontdimen4\font\relax}
\providecommand\BIBforeignlanguage[2]{{%
\expandafter\ifx\csname l@#1\endcsname\relax
\typeout{** WARNING: IEEEtran.bst: No hyphenation pattern has been}%
\typeout{** loaded for the language `#1'. Using the pattern for}%
\typeout{** the default language instead.}%
\else
\language=\csname l@#1\endcsname
\fi
#2}}

\bibitem{mdeComplex}
\BIBentryALTinterwordspacing
J.~B{\'e}zivin, R.~F. Paige, U.~A{\"s}mann, B.~Rumpe, and D.~C. Schmidt,
  ``{Perspectives Workshop: Model Engineering of Complex Systems (MECS)},''
  Dagstuhl Seminar Proceedings, 2008. [Online]. Available:
  \url{http://drops.dagstuhl.de/opus/volltexte/2008/1604}
\BIBentrySTDinterwordspacing

\bibitem{artist2}
ARTIST, ``Network of excellence on embedded system design,'' 2010,
  \url{http://www.artist-embedded.org/}, visited on June 15th 2010.

\bibitem{isorc09}
ISORC09, ``{12th IEEE Int. Symp. on Object/Component/Service-Oriented Real-Time
  Distributed Computing},'' 2009,
  \url{http://www.dcl.info.waseda.ac.jp/isorc2009/}, visited on June 15th 2010.

\bibitem{autosar}
AUTOSAR, ``Automotive open system architecture,'' 2010,
  \url{http://www.autosar.org/}, visited on June 15th 2010.

\bibitem{rtdescribe}
RT-Describe, ``{Iterative Design Process for Self-Describing Real Time Embedded
  Software Components},'' 2010,
  \url{http://www.esk.fraunhofer.de/EN/press/pm0911RTDescribe.jsp}, visited on
  June 15th 2010.

\bibitem{omgUml}
{OMG UML}, ``{Unified Modeling Language (UML) Superstructure specification
  v2.2, formal/09-02-02},'' February 2009.

\bibitem{omgSysml}
{OMG SysML}, ``{OMG Systems Modeling Language (SysML) specification v1.1,
  formal/2008-11-02},'' November 2008.

\bibitem{omgMarte}
{OMG MARTE}, ``{A UML Profile for MARTE: Modeling and Analysis of Real-Time
  Embedded systems, Beta 2, ptc/2008-06-08},'' June 2008.

\bibitem{rosta}
\BIBentryALTinterwordspacing
{Robot Standards and Reference Architectures (RoSTa)}, ``{Coordination Action
  funded under the European Union’s Sixth Framework Programme (FP6)},''
  February 2010. [Online]. Available:
  \url{http://wiki.robot-standards.org/index.php/Middleware}
\BIBentrySTDinterwordspacing

\bibitem{player}
B.~Gerkey, R.~T. Vaughan, and A.~Howard, ``The player/stage project: Tools for
  multi-robot and distributed sensor systems,'' in \emph{Proc. of the 11th Int.
  Conf. on Advanced Robotics (ICAR)}, Coimbra, Portugal, June 2003, pp.
  317--323.

\bibitem{ros}
{ROS}, ``Robot operating system,'' 2010, \url{http://www.ros.org/}, visited on
  June 15th 2010.

\bibitem{msrs}
Microsoft, ``{Microsoft Robotics Developer Studio},'' 2010,
  \url{http://msdn.microsoft.com/en-us/robotics/default.aspx}, visited on June
  15th 2010.

\bibitem{omgdds}
{OMG DDS}, ``{Data Distribution Service for Real-time Systems (DDS) v1.2,
  formal/2007-01-01},'' January 2007.

\bibitem{brugali2009}
D.~Brugali and P.~Scandurra, ``{Component-Based Robotic Engineering (Part
  I)},'' \emph{IEEE Robotics \& Automation Magazine}, vol.~16, no.~4, pp.
  84--96, Dezember 2009.

\bibitem{brugali2010}
D.~Brugali and A.~Shakhimardanov, ``{Component-Based Robotic Engineering (Part
  II)},'' \emph{IEEE Robotics \& Automation Magazine}, vol.~17, no.~1, pp.
  100--112, March 2010.

\bibitem{brics-ICR2010}
R.~Bischoff, T.~Guhl, E.~Prassler, W.~Nowak, G.~Kraetzschmar, H.~Bruyninckx,
  P.~Soetens, M.~Haegele, A.~Pott, P.~Breedveld, J.~Broenink, D.~Brugali, and
  N.~Tomatis, ``{BRICS -- Best practice in robotics},'' in \emph{Proc. of the
  Joint 41st International Symposium on Robotics and the 6th German Conference
  on Robotics}, 2010, pp. 968--975.

\bibitem{orocos03}
H.~Bruyninckx, P.~Soetens, and B.~Koninckx, ``{The real-time motion control
  core of the Orocos project},'' in \emph{{Proc. of the IEEE Int. Conf. on
  Robotics and Automation (ICRA)}}, vol.~2, 2003, pp. 2766--2771.

\bibitem{anthony2010}
A.~Mallet, C.~Pasteur, M.~Herrb, S.~Lemaignan, and F.~Ingrand, ``{GenoM3:
  Building middleware-independent robotic components},'' in \emph{Proc. of the
  IEEE Int. Conf. on Robotics and Automation (ICRA)}, 2010.

\bibitem{omg-robotics}
{OMG Robotics}, ``{OMG Robotics Domain Task Force},'' 2010,
  \url{http://robotics.omg.org/}, visited on June 15th 2010.

\bibitem{joserV3CMM}
D.~Alonso, C.~Vicente-Chicote, F.~Ortiz, J.~Pastor, and B.~\'{A}lvarez,
  ``{V$^{3}$CMM: a 3-View Component Meta-Model for Model-Driven Robotic
  Software Development},'' \emph{Journal of Software Engineering for Robotics},
  vol.~1, no. January, pp. 3--17, 2010.

\bibitem{generationGap}
J.~Vlissides, ``{Pattern Hatching -- Generation Gap Pattern},''
  \url{http://researchweb.watson.ibm.com/designpatterns/pubs/gg.html}, visited
  on June 15th 2010.

\bibitem{schlegel2006}
C.~Schlegel, ``{Communication Patterns as Key Towards Component-Based
  Robotics},'' \emph{Int. Journal of Advanced Robotic Systems}, vol.~3, no.~1,
  pp. 49 -- 54, 2006.

\bibitem{Schlegel2009}
C.~Schlegel, T.~Ha{\ss}ler, A.~Lotz, and A.~Steck, ``Robotic software systems:
  From code-driven to model-driven designs,'' in \emph{International Conference
  on Advanced Robotics (ICAR)}, June 2009.

\bibitem{smartsoft-sourceforge}
SmartSoft, \url{http://smart-robotics.sf.net/}, visited on June 15th 2010.

\bibitem{ace}
D.~Schmidt, ``{The ADAPTIVE Communication Environment},''
  \url{http://www.cs.wustl.edu/~schmidt/ACE.html}, visited on June 15th 2010.

\bibitem{schlegel-diss}
C.~Schlegel, ``Navigation and execution for mobile robots in dynamic
  environments -- an integrated approach,'' Ph.D. dissertation, Faculty of
  Computer Science, University of Ulm, 2004.

\bibitem{eclipse}
E.~M. Project, \url{http://www.eclipse.org/modeling/}, visited on June 15th
  2010.

\bibitem{cheddar}
CHEDDAR, ``The cheddar project,''
  \url{http://beru.univ-brest.fr/~singhoff/cheddar/}, visited on June 15th
  2010.

\end{thebibliography}

\end{document}